# PersianRAG: A Retrieval-Augmented Generation System for Persian Language

Hossein Hosseini
Department of software engineering,
University of Isfahan,
Isfahan, Iran
h.hosseini@eng.ui.ac.ir

Mohammad Sobhan Zare
Department of software engineering,
University of Isfahan,
Isfahan, Iran
s.zare@eng.ui.ac.ir

Amir Hossein Mohammadi
Department of software engineering,
University of Isfahan,
Isfahan, Iran
a.mohammadighahderijani@eng.ui.ac.ir

Arefeh Kazemi
ADAPT Research Centre,
Dublin City University,
Dublin, Ireland
arefeh.kazemi@dcu.ie

Zahra Zojaji
Department of software engineering,
Faculty of Computer Engineering,
University of Isfahan,
Isfahan, Iran
z.zojaji@eng.ui.ac.ir

Mohammad Ali Nematbakhsh
Department of software engineering,
Faculty of Computer Engineering,
University of Isfahan,
Isfahan, Iran
Nematbakhsh@eng.ui.ac.ir

*Abstract*— Retrieval augmented generation (RAG) models, which integrate large-scale pre-trained generative models with external retrieval mechanisms, have shown significant success in various natural language processing (NLP) tasks. However, applying RAG models in Persian language as a low-resource language, poses distinct challenges. These challenges primarily involve the preprocessing, embedding, retrieval, prompt construction, language modeling, and response evaluation of the system. In this paper, we address the challenges towards implementing a real-world RAG system for Persian language called PersianRAG. We propose novel solutions to overcome these obstacles and evaluate our approach using several Persian benchmark datasets. Our experimental results demonstrate the capability of the PersianRAG framework to enhance question answering task in Persian.

*Keywords—Retrieval Augmented Generation, Large Language Models, , Persian, PersianRAG.*

## I. Introduction

In recent years, the field of NLP has witnessed significant advancements, especially in the development of LLMs capable of generating coherent and contextually relevant text. Despite their impressive capabilities, generative language models often tend to provide outdated information or fabricate facts a phenomenon commonly referred to as "Hallucination". This limitation persists even when models are aligned with human preferences through reinforcement learning [1] or style alignment techniques [1-4]. RAG systems have emerged as a promising solution to these challenges. By integrating the strengths of pre-trained models and retrieval mechanisms, RAG provides a powerful framework that enhances model performance and reduces errors in generated content [5]. Additionally, RAG enables the rapid deployment of applications tailored to specific organizations and domains without needing to update the underlying model's parameters, provided that relevant documents are available for retrieval. Several methods have been proposed to enhance LLMs through query-dependent retrieval [5-7]. A typical RAG process includes critical components such as embedding (semantic representation of documents and queries), retrieval (efficient access to relevant documents), and generation (producing responses based on retrieved information). Implementing RAG requires key decisions regarding document segmentation, selecting embedding models for semantic representation, choosing vector databases for efficient storage and retrieval, and optimizing large language models (refer to Figure 1). The inherent complexity at each stage of this process creates significant challenges in implementing RAG systems which are more highlighted in a low resource language like Persian. One approach involves using embedding models directly to calculate semantic similarities between queries and documents. These embedding models are often trained adversarially using positive and negative query-response pairs [8, 9]. The choice and combination of techniques at each stage profoundly affect the effectiveness and efficiency of RAG systems. Furthermore, according to the latest reviews, there no research on optimizing RAG implementations for Persian languages, with a focus on core components such as retrieval, embedding, and generative models. Historically, most NLP research and development have focused on English, resulting in a shortage of resources, tools, and datasets for other languages. Consequently, the development of advanced NLP systems for languages like Persian lags behind the growing need for such technologies.

This paper provides the PersianRAG framework as a real implementation of Persian RAG system through addressing several challenges. The various stages of developing the PersianRAG system includes, data preprocessing, embedding, retrieval, generation and hyperparameter optimization. In proposing PersianRAG, we introduce innovative approaches tailored to the Persian language. These approaches include using advanced tools for Persian text processing, optimizing document segmentation strategies to accommodate Persian language characteristics, selecting and fine-tuning appropriate embedding models for semantic representation in both unstructured text and structured data such as tables, and designing effective strategies to reduce hallucinations in generated responses. Moreover, we emphasize the importance of rigorous system performance evaluation using Persian-specific datasets to ensure the reliability and accuracy of the PersianRAG system with several datasets introduced in this research.

## II. Retrieval Augmented generation

RAG helps produce more accurate and comprehensive responses by combining two approaches retrieval and generation. Unlike purely generative models that respond solely based on their training data, RAG can refer to external knowledge bases and retrieve real-time, accurate information to enhance its responses. In this architecture:

- Information retrieval is a process that extracts relevant external knowledge related to the user's query, rather than focusing on information that the language model can easily understand. It aims to find the most relevant pieces of knowledge to support the model's response generation.

- The RAG approach, which combines retrieved information with the language model's ability to understand and generate human-like text, leads to more accurate and relevant results. This method is particularly useful in domains requiring deep and up-to-date knowledge, such as medicine, law, and science, where it can improve response quality and reduce errors in language models.

This reception from RAG systems was primarily due to two factors: first, the impressive capabilities offered by LLMs, and second, because it combines the strengths of retrieval-based and generation-based models to improve text generation tasks. Essentially, the advantage of using RAG lies in leveraging the power of LLMs to generate responses based on documents that these LLMs may not have previously seen or been trained on. In specific domains, this means obtaining high-quality and precise answers to questions for which an LLM might not have an inherent response.

## III. PersianRAG Architecture

Despite major advancements in the capabilities of language models, RAG remains a highly relevant and efficient option for generating accurate and enriched responses by incorporating information from external sources. Despite the proven effectiveness of RAG in question-answering systems, multilingual RAG and specifically RAG in Persian, a language with unique characteristics and resource limitations has received less emphasis in research, despite its importance and widespread application. Nevertheless, recent studies have begun to address this gap. Therefore, the current study presents the first successful implementation of Persian RAG system called PersianRAG. The architecture of the PersianRAG system is depicted in Fig. 1.

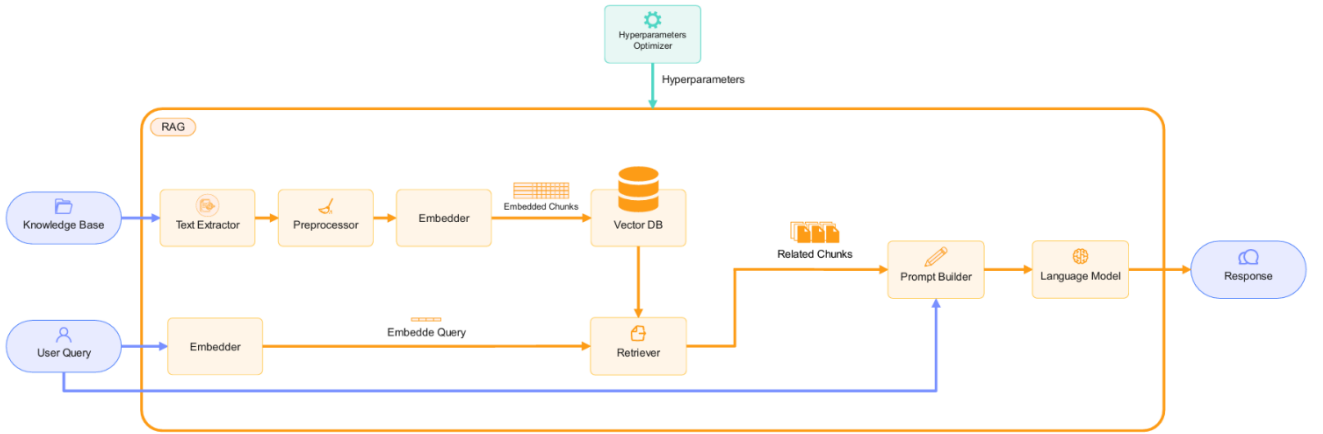

Fig 1. RAG pipeline architecture

The components of this architecture are described as follows.

### A. Knowledge Base

The knowledge base serves as a repository of structured information, consisting of processed data that has been organized into meaningful content. This content may include a collection of various texts, scientific articles, wikis, organizational documents, and other curated information sources. The knowledge base is continuously updated and serves as the primary input for information retrieval in the RAG system.

### B. Text Extractor

The text extractor is responsible for processing and extracting raw textual content from the knowledge base. These contents may include structured or unstructured documents that need to be transformed into suitable formats for preprocessing.

### C. Preprocessor

After extracting information from the knowledge base, this component cleans and optimizes the texts. Preprocessing includes activities such as removing noise, additional markings, and certain specific characters in various languages. For Persian, for example, this might involve removing half-spaces. This stage is crucial, as the quality of input data has a direct impact on final results.

### D. Embedder

The embedder plays a key role in converting texts into vector representations. This process involves using deep learning models to compress textual information into numerical vectors. These vectors, although more compact and with fewer dimensions, preserve the semantic information of the texts and convert them into an efficient vector representation for further processing.

*E. Vector Database*

Once the texts have been converted to vectors, they are stored in a vector database. This database is optimized to facilitate fast and efficient retrieval of similar vectors. Using a vector database allows the system to perform rapid searches based on textual vectors. One of the leading vector databases, which was also used in this research, is Elasticsearch.

*F. User Query*

This section relates to the user's input. The user presents a query or request to the system, which could be a simple or complex text that requires processing by the language model.

*G. Embedder (For user query)*

As in the previous stage, the user's query is also transformed into a numerical vector by an embedder. This vector represents the semantic meaning of the user's query and is used to compare it with the stored vectors in the Vector Database.

*H. Retriever*

The retriever compares the user's query vector with the vectors stored in the vector database to find the sections of the knowledge base that have the highest semantic similarity to the query. This optimized retrieval process allows the system to pass precise information to subsequent stages.

*I. Related Chunks*

This section collects texts identified by the retriever as relevant and feeds them into the system as the next input. These are key parts of information directly related to the user's query.

*J. Prompt Builder*

The Prompt Builder combines the user's query with relevant sections extracted from the knowledge base, forming a complete and comprehensive prompt for the language model. The combination ensures that the language model receives relevant information alongside the user's question, leading to more accurate responses.

*K. Language Model*

The language model is responsible for generating the final response. Utilizing the final query constructed by the Prompt Builder, this model produces a relevant and logical answer. Language models, leveraging millions of parameters and based on extensive textual data on which they have been trained, possess the ability to generate natural and human-like language. In this research, we have employed both extractive models, specifically the BERT-base models, and their generative counterparts, namely LLMs.

*L. Response*

The response generated by the language model is presented to the user. This response may include textual explanations, numerical results, or even advanced analyses.

*M. Hyperparameters Optimizer*

The hyperparameters optimizer is a component of the system responsible for enhancing its overall performance. By adjusting various hyperparameters at different stages, such as data processing, vector transformations, and information retrieval, this component helps the system operate at maximum efficiency. These optimizations can include parameters like the Top K, which refers to the number of most relevant and similar results retrieved from the documents, embedding dimensions, the number of retrieved chunks, chunk size, chunk overlap, and other key variables.

IV. RELATED WORKS

Recently, the concept of RAG, as a hybrid approach that integrates information retrieval with natural language generation, has garnered significant attention. In 2020, Lewis et al. [1] introduced this concept. Introduced the RAG framework to address the limited capabilities of pre-trained language models and to enhance the model's ability to generate informed and relevant responses, which was remarkably well-received. Since there is little or no work in implementing RAG for Persian, in this section we review similar research in Non-English languages.

For example, the study presented in [4] discusses the application of RAG in multilingual settings, focusing specifically on improving the performance of RAG models when working with non-English languages. In the research by Nadezhda Chirkova et al., the importance of making specific configurations and considering particular components necessary for multilingual RAG is addressed. The article mentions that deploying high-quality RAG in multilingual settings requires access to robust multilingual retrievers and generators, as well as high-quality multilingual evaluation. The generative component must be able to produce text fluently and correctly in the user's language, while also understanding documents in different languages and following instructions specified in the prompt. While recent advancements in NLP and information retrieval have made suitable candidate components available, this article introduces a multilingual RAG pipeline that claims to have not been previously evaluated in the literature. Therefore, it can serve as an appropriate starting point for designing a Persian RAG, which is considered a multilingual RAG. Although this article pays good attention to various parts of a RAG system for multilingual implementation, addressing Persian RAG requires research tailored to the script, grammar, and vocabulary of the Persian language to achieve the best results.

Also, the article by Mohammad A. Abdullah [5] presents a case study on the implementation and evaluation of RAG for Arabic texts. This paper focuses exclusively on exploring various semantic embedding models in the retrieval phase and several LLMs in the generation phase to determine which LLMs and semantic embedding models are suitable for the Arabic language and its unique characteristics. Two objectives are pursued in this study. One of the primary goals is to evaluate the retrieval aspect of RAG by examining how different embedding models perform in the context of Arabic. As the second goal, the paper conducts preliminary analyses to investigate which LLM performs better as a generator in Arabic (with a focus on open-source models).

In another study conducted by Hazem Abdelazim et al. [6] on the Arabic language, it is noted that a critical element in RAG systems is the information retrieval component, at the core of which lies the vector embedding process known as "semantic embedding." Their research is focused on studying a set of multilingual semantic embedding models with the aim of effectively enhancing the model's ability to understand and generate Arabic text. This study, conducted using ARCD (the publicly available Arabic Reading Comprehension Dataset), has examined various embedding models, including OpenAI's Ada, Google's Language-Agnostic Sentence Embedding

BERT (LaBSE) [10], Cohere Embed-Multilingual-v3.0 [11], MPNet [8], versions one and two of HuggingFace's DistilBERT [12], Meta's SONAR [13] and Microsoft's E5 embedding models [14].

Ultimately, the results of this research indicated that Microsoft's E5 sentence embedding model outperformed all other models on the ARCD dataset, achieving a Recall@10 of over 90%. This study has paid special attention to the embedding process in multilingual RAG (in this case, Arabic) and provides valuable insights into the process of selecting the information retrieval component for multilingual RAG systems, particularly for languages that share the same script as Persian. In this paper, the authors did not use a vector database but opted for the direct use of cosine similarity to match query embeddings against document embeddings. While this approach may slow down the matching process in a real-world environment, it should have little impact on the research findings presented in the paper. Although this article does not, by any means, represent a complete RAG system, it offers a clear pathway for understanding semantic embedding models in designing a multilingual RAG system, and particularly RAG in the Persian language, which shares the same script as Arabic.

As the last work reviewed in the Arabic language, we can refer to the research conducted by Ali Mahbub et al. [15]. The paper, focusing on semantic search in Arabic, strives to create a simple yet robust benchmark for semantic search in the Arabic language. The importance of semantic search is emphasized in this article, noting that semantic search interprets the meanings and relationships between words with the aim of mimicking human understanding. This enhances user experience in various applications, including web search engines, personalized content recommendation systems, and also RAG. The article further discusses that by combining a sophisticated retrieval mechanism with a powerful generation model, RAG systems can produce accurate and contextually relevant responses, significantly improving the limitations of standalone language models in terms of accuracy and human-like text generation. Integrating semantic search into RAG systems is crucial, especially for processing complex queries or those requiring deep contextual understanding, making it a cornerstone for enhancing retrieval accuracy and the quality of generated content. However, similar to other reviewed works and like most research efforts in NLP tasks, Arabic semantic search and RAG lag behind other languages due to challenges inherent in the Arabic language, including complex morphology, the diversity of its dialects, and the scarcity of datasets [16, 17]. Ultimately, the goal of this paper was to evaluate the effectiveness of semantic search in Arabic language processing alongside its impact on the performance of RAG systems specifically designed for Arabic question-answering use cases.

Other research has also been conducted on the effects of different components of RAG systems. For instance, the article by Yuanjie Liu et al. [18] introduces a comprehensive Chinese benchmark for RAG in large language models . This paper addresses the challenge of evaluating RAG systems by highlighting that most benchmarks primarily focus on question-answering applications and tend to overlook other potential scenarios where RAG could be beneficial. To overcome this limitation, they develop a large-scale and more comprehensive benchmark that evaluates all components of RAG systems across various RAG application scenarios.

Specifically, the paper describes the interactions between users and knowledge bases in a RAG system in terms of CRUD actions and categorizes RAG applications into four distinct types: Create, Read, Update, and Delete (CRUD). The 'Create' category refers to scenarios requiring the generation of original and diverse content. 'Read' involves answering complex questions in knowledge-intensive situations. 'Update' focuses on revising and correcting inaccuracies or inconsistencies in existing texts. Lastly, 'Delete' pertains to the task of summarizing extensive texts into more concise forms.

For each of these CRUD categories, the study has created and utilized different datasets to evaluate the performance of RAG systems. Additionally, the effects of various RAG system components such as the retriever, context length, knowledge base construction, and the LLM have been analyzed. Ultimately, this research provides valuable insights for optimizing RAG technology for different scenarios, offering a comprehensive evaluation framework that contributes to the advancement of RAG systems.

However, this is not the only work conducted in the field of multilingual RAG focused on the Chinese language. In another article presented by Shuting Wang et al. [19], it is noted that existing studies, such as the one by Chen et al. [20], primarily rely on general knowledge sources like Wikipedia as external knowledge bases for evaluating RAG models [21], [22, 23]. The authors argue that such an approach may not fully assess the capability of RAG models to solve domain-specific problems. Therefore, utilizing domain-specific datasets and questions is essential for evaluating the ability of LLMs to effectively leverage external knowledge in specific contexts. In this paper, LLMs configured with RAG are evaluated within a specific domain context college enrollment. The study identifies six essential capabilities required for RAG models: conversational RAG, analysis of structured information, fidelity to external knowledge, noise reduction, solving time-sensitive issues, and understanding multi-document interactions. For each capability, relevant datasets with shared tasks are considered to assess the performance of RAG models.

The paper evaluates popular LLMs, including models like Llama, Baichuan, ChatGLM, and GPT. Experimental results demonstrate that existing closed-book LLMs struggle with domain-specific questions, highlighting the need for RAG systems to address these challenges. Furthermore, there is room for RAG models to enhance their abilities in understanding conversational history, analyzing structured information, reducing noise, processing multi-document interactions, and maintaining fidelity to specialized knowledge.

As previously noted, unfortunately, most efforts in this field have focused on English as the natural language of input data (including the language of user queries and the stored knowledge) in their experiments. In other cases reviewed as related works in this section, solutions for implementing multilingual RAG have been proposed, and the importance of parameter tuning and prompt engineering in multilingual RAG is evident. Additionally, the implementation of RAG for the Arabic language particularly the embedding process, given that it shares the same script as Persian has been examined, which can illuminate the path for the research process of Persian RAG, considering the shared script with Arabic.

## V. CHALLENGES

### A. Reading Persian PDFs without character distortion

Reading Persian PDFs presents a major challenge for RAG systems. This challenge arises particularly due to the presence of half-space characters and the Persian 'ى', as well as specific encodings used in certain organizations. As a result, when extracting text from Persian PDFs, the text might become completely scrambled, or characters might incorrectly stick together or be separated. For example, the character "ه" might be written separately, or "ى" may display incorrectly. These problems significantly hinder the comprehension of text by large language models, and the more non-standard or problematic data there is, the lower the accuracy of the entire pipeline.

Additionally, some organizations or companies maintain their PDFs as scanned copies with no access to the original Text version. For these types of files, advanced image-to-text conversion models and various (Optical Character Recognition) OCR tools were tested.

Initial experiments indicated that the best method for extracting text from these files is through OCR. In this regard, open-source multilingual OCR tools with Persian recognition capabilities that operate offline were evaluated. The tools tested included Tesseract OCR, ABBYY FineReader, py2pdf, and Easy OCR. The results showed that the best tool for offline OCR is Google's Tesseract OCR. Although Tesseract performs better in recognizing Persian text compared to similar tools, it still faces challenges, particularly in correctly reading numbers in the text.

### B. Finding the best embedding model or approach for the Persian language

Finding the best embedding model or approach for the target language in RAG systems is of great importance. These models convert text into numerical vectors, which are semantic representations of the text used for information search and retrieval. The quality and accuracy of information retrieval heavily depend on the performance of the embedding model.

To address this challenge, the following methods can be utilized:

- **Overall Performance Comparison:** Evaluating different embedder tools using metrics like retrieval accuracy, processing speed, and resource consumption helps identify the tools that perform best.
- **Domain-Specific Testing:** Testing embedders with data relevant to the specific domain ensures they perform well in the desired field, ensuring optimal real-world performance.
- **Language Support Evaluation:** Ensuring the chosen embedder has strong support for the target language, especially for low-resource languages like Persian, is crucial.
- **Scalability Assessment:** Checking the embedder's ability to handle large volumes of data and perform well at scale is essential for industrial and commercial applications where text similarity is high.
- **Semantic Testing:** Conducting tests to ensure the embedder maintains complex semantic relationships in the target language helps guarantee accurate semantic retrieval.

These methods can help select the best embedder for low-resource languages like Persian and improve the performance of RAG systems in these areas.

Based on evaluations of Persian as the target language, over 40 embedding models, thought to be capable of embedding Persian text, were assessed. Some of them were eliminated in the early stages due to their poor results, and the remaining models were further examined. Table I presents these evaluation results, highlighting the performance of different embedders for Persian text processing.

TABLE I. SUMMARY OF EMBEDDING MODELS' RESULTS

| Model | Top 1 | Top 2 | Top 3 | Missed |
|---|---|---|---|---|
| paraphrase-multilingual-mpnet-base-v2 | 67.2 | 10.8 | 5.4 | 3.5 |
| bert-base-parsbert-uncased | 69.6 | 13.3 | 3.9 | 4.6 |
| distiluse-base-multilingual-cased-v2 | 70.2 | 11 | 4.2 | 5.3 |
| dpr-xm | 71.4 | 10.4 | 4.6 | 3.5 |
| AviLaBSE | 79.3 | 7.9 | 3.6 | 3.7 |
| sentence-embedding-LaBSE | 79.3 | 7.9 | 3.6 | 3.7 |
| LaBSE | 80.7 | 9.1 | 2.5 | 2.3 |
| LaBSE-sentence-embeddings | 80.7 | 9.1 | 2.5 | 2.3 |
| persian-sentence-transformer-news-wiki-pairs-v3 | 88.1 | 5.3 | 2.1 | 1.3 |
| jina-embeddings-v3 | 88.8 | 4.1 | 1.7 | 1.9 |
| cohere-embed-multilingual-v3.0 | 93.5 | 2.9 | 1.1 | 0.3 |

Most models reviewed were open-source and available on platforms like Hugging Face, except for one from Cohere, which is not open-source. This diversity allowed us to comprehensively assess the performance of various models in Persian language processing.

To compare different embedder models, a subset of data was extracted from the main dataset, PersianQuad [25]. The original dataset consists of columns for Paragraph, Question, and GoldAnswer, designed for training and evaluating QA systems. However, for the purpose of comparing embedders, only the Paragraph and Question columns were used. The evaluation was conducted within the context of an information retrieval framework, where better embeddings result in improved retrieval performance.

Next, this subset was processed separately by each embedder model, and the results were stored in a vector database. Using Cosine Similarity, the quality of various embeddings was evaluated by selecting the Question as input and searching for the most similar paragraphs in the vector database. The top ten most similar paragraphs were compared to the actual paragraph, and the rank of the actual paragraph among the top ten was recorded. If the rank was 'n', a unit was added to Top n. For example, if in 1,000 tuples, the actual

paragraph was ranked first 748 times, the Top 1 value for that embedder would be 748

We concluded that combining different retrievers and using a hybrid approach could increase pipeline accuracy by up to 4%. This approach includes using search-based retrievers like BM25 along with dense retrievers. By adjusting various top_k values and adding a document_joiner at the end of the indexing pipeline, better accuracy could be achieved compared to using either method alone [26]. In our final pipeline, the top_k value for BM25 was set to 4 and for dense retriever to 8. The document_joiner directs all documents to extractive or generative models, sending up to 12 documents to the model. It also removes duplicates and considers the document with the highest score.

And also incorporating Reranker models into RAG systems significantly enhances retrieval accuracy by refining the ranking of retrieved chunks. Our experiments demonstrate that after retrieving results from a vector database using the 'cohere-embed-multilingual-v3.0' embedding model, the addition of the 'cohere-rerank-multilingual-v3.0' as a Reranker substantially improves performance. Specifically, the Top-1 accuracy of correctly indexed chunks increased from 93.5% to 98.7%. This improvement indicates that the Reranker effectively reorders the retrieved chunks, ensuring that the most relevant chunks are ranked first, which enhances the overall effectiveness of the RAG system's retriever module.

*C. Prompt engineering*

In this research, focusing on the Persian language, we concluded that selecting the appropriate language for writing prompts based on different LLMs has a significant impact on the overall performance of the system. Language models that support only a limited number of languages react differently to the input prompt language compared to multilingual models. Specifically, in monolingual models or those with limited support, using a language in which the model has greater proficiency (usually English) leads to a better understanding of the instructions and assigned tasks. In contrast, in multilingual models like aya and command-r, which were examined in this study, using English as the prompt language can cause confusion, and sometimes appearance of the words from the prompt language in the output. For example, even when emphasizing the production of the answer in Persian, English words may appear among Persian sentences. Therefore, for multilingual models, writing prompts in the target language (Persian) can help produce smoother outputs and reduce the generation of irrelevant words. In this study, changing the prompt language from English to Persian led to a significant reduction in the occurrence of English words in the output, indicating the importance of choosing the appropriate language for prompts in multilingual models.

- **Separation of Prompt Components**: The investigation focused on using specific markers to clearly separate different components of the prompt, such as the instructions, retrieved results, and user query. Generally, common separators that language models have been trained on were used, following templates like:

```
### Instructions
[Specific guidelines for the AI model]

### User Query
[The user's question or request]

### Retrieved Information
[Relevant data retrieved by the RAG system]

### Your Response:
[Space for the AI-generated response]
```

Can be significantly helpful and assist the language model in better understanding. Additionally, utilizing Markdown symbols to separate different parts of the text, rows and columns in tables, and optimizing table styles, according to our team's investigations, results in receiving more accurate and better responses from LLMs.

- **Adding Metadata:** An effective strategy to improve the performance of RAG systems is to include brief contextual information before each retrieved result. This approach helps LLMs better comprehend the relationship between the retrieved results and the user's query, leading to more accurate responses. In real operational environments, which encompass a vast array of diverse documents such as sales reports and meeting minutes, adding metadata can significantly prevent semantic errors. For instance, in response to a query like "Sales report of the section under discussion following the June meeting", without metadata, the retrieval module might return all results related to June without considering the relevant year. This could result in the LLM generating irrelevant and misleading answers. To resolve this issue, appropriate metadata, such as precise temporal information (e.g., datetime), can be added to the user's query, substantially increasing the accuracy and correctness of the responses. This approach assists both the LLM and the retrieval module in understanding the user's exact intent when referencing a specific time frame. It is important to note that the type and number of required metadata should be adjusted according to the application domain and the general pattern of expected queries in the RAG system. In some cases, merely adding temporal information may not suffice, and it may be necessary to incorporate additional contextual information. To optimize this process, another LLM chain can be employed at the beginning of the pipeline to intelligently identify and add the necessary types and amounts of metadata. In conclusion, by implementing the aforementioned set of strategies and conducting continuous optimizations, we were able to significantly enhance the performance of the implemented RAG system in terms of information retrieval and generating relevant responses. These advancements are particularly important for low-resource languages like Persian, which face specific challenges in natural language processing.

- **Language Model:** The language model is responsible for generating the final response. Utilizing the final query constructed by the Prompt Builder, this model produces a relevant and logical answer. Language models, leveraging millions of parameters and based on extensive textual data on which they have been trained, possess the ability to generate natural and human-like language. In this research, we have employed both extractive models, specifically the BERT-base models, and their generative counterparts, namely LLMs.

*D. Finding a Model for Answering Questions in the Target Language*

In the present study, the main challenge was selecting an appropriate model for answering questions in Persian using RAG technology. This challenge arises because most existing models are optimized for the English language and may not perform desirably in Persian, especially given the limitations of training resources. To address this issue, a comprehensive process for model evaluation and selection was designed and implemented.

Table VI presents the results of evaluation of LLMs for the PersianRAG.

TABLE II. RESULTS OF EVALUATING LANGUAGE MODELS IN THE IMPLEMENTED RAG PIPELINE

| Model | Wrong | Middle | Correct |
| --- | --- | --- | --- |
| GPT 3 | 48% | 12% | 40% |
| PXRL | 18% | 2% | 80% |
| Aya 101 | 6% | 3% | 91% |

Ultimately, two models with the best performance were identified:

- **Persian-XLM-Roberta-Large Model:** An encoder-only model that enables the generation of extractive responses.
- **Command-R Model:** This model was selected due to its architectural similarity to the Aya-101 model, which had previously yielded satisfactory results. A real RAG pipeline was developed based on this model, and through various versions, its accuracy gradually improved.

TABLE III. ACCURACY ACHIEVED WITH THE COMMAND-R MODEL ON THE PERSIANQUAD DATASET

| Pipeline | Wrong | Middle | Correct |
| --- | --- | --- | --- |
| v1.0 | 24% | 0% | 76% |
| v2.0 | 13% | 0% | 87% |
| v3.0 | 11% | 0% | 89% |

Ultimately, the highest accuracy achieved in the initial version was 76%, obtained using the Persian-XLM-RoBERTa-Large model. By implementing improvements in embeddings, hybridizing the retriever, adjusting the chunk size, and other optimizations, this accuracy increased to 89% on 1,000 test samples from the PersianQuAD dataset.

*E. Selecting the Best Fine-Tuned LLMs for the Target Language*

In the pursuit of enhancing the performance of LLMs for low-resource languages, fine-tuning these models for the target language can be an effective strategy. However, selecting the best method and optimized model presents a significant challenge in this domain. This research addresses this challenge in the context of the Persian language.

In the present study, the performance of several fine-tuned models was evaluated. These models included PersianMind-7B [29] and PersianLlama-7B [30] from the University of Tehran (based on Llama 2), and NeuraOrcaGemma-7B (based on Google's Gemma model). The results of these evaluations indicated that none of these models achieved acceptable accuracy and often suffered from hallucination errors

Deeper analyses revealed that some language models are fundamentally unsuitable for fine-tuning on a specific target language. This issue stems from the volume of training data. According to our findings, at least 0.02% of the model's total training corpus must be dedicated to the target language to achieve desirable performance; otherwise, the weight update process leads to incomplete learning of associations and meanings in the new language [31].

Furthermore, it was observed that fine-tuning models for the Persian language not only results in insufficient understanding of this language but also adversely affects the model's performance in other languages, including the model's primary language (English). For instance, the PersianLlama-7B model (based on Llama 2) and the NeuraOrcaGemma-7B model (based on Gemma), which were optimized to improve understanding of Persian, exhibited a significant decline in their performance accuracy in English.

These findings underscore the importance of considering the volume and quality of training data in the fine-tuning process of large language models for low-resource languages. They also highlight the need to develop new and more efficient methods for optimizing these models for Persian.

*F. Precise Evaluation of the RAG System's Performance in Answering Factoid Questions*

In the process of developing and optimizing RAG systems for the Persian language, one of the fundamental challenges is the lack of sufficient insight into the model's performance in responding to various types of factoid questions. This issue can lead to inefficient system design and evaluation.

Factoid questions play a crucial role in assessing the quality and efficiency of RAG systems due to their high accuracy requirements and specific answers that are usually directly extractable from the text. These types of questions, especially in low-resource languages like Persian, serve as appropriate benchmarks for measuring the system's ability to retrieve and provide precise information. In this study, the evaluation process was designed and implemented with a special focus on factoid questions in Persian, considering the generalizability of the RAG system across different domains. This approach allows for a more precise and comprehensive assessment of the system's performance in processing and responding to fact-based questions in Persian. This revision integrates the importance of factoid questions into the challenge description, addressing the editor's concern about the paragraph structure.

*G. Lack of Factoid Datasets for Evaluating RAG Systems*

One of the fundamental challenges in developing RAG systems for low-resource languages, including Persian, is the lack of appropriate factoid datasets and sufficient evaluation mechanisms. This deficiency makes it challenging to accurately assess the model's performance in generating correct and relevant responses. In this research, considering a gradual approach in system development from a general state towards a specific domain (documents and educational data) datasets appropriate for each stage have been designed and employed.

To address this challenge and gain a more precise understanding of the RAG system's performance when handling factoid questions in Persian, two key datasets have been developed and utilized:

**A new dataset Derived from PersianQuAD:** This dataset comprises unique paragraphs extracted from the PersianQuAD dataset. These paragraphs are presented to the RAG system as a collection of documents, and the system's performance is evaluated based on the questions and answers available in PersianQuAD. This approach enables the assessment of the system's efficiency in responding to factoid questions in Persian within the realm of general inquiries. The structure of this dataset includes columns for the paragraph, expected answer, question, and title.

IV. STRUCTURE OF THE PERSIANQUAD DATASET TABLE

| paragraph | question | gold_answer |
|---|---|---|
| Paragraph containing the question and answer | Question posed to the model | The expected answer |

**PorsemanQuad Dataset:** This dataset, extracted from a collection of regulations and guidelines from the University of Isfahan, consists of 50 pairs of questions and answers. The structure of this dataset includes columns for the question, answer, related paragraph, question type (factoid from a table or plain text), and the document name. Designed to evaluate the RAG system's performance when dealing with university regulations in Persian, this dataset enables a more focused assessment of factoid questions. The responses generated by the system are added to a separate column in this dataset and are evaluated.

TABLE V. STRUCTURE OF THE PORSEMANQUAD DATASET

| source file | question type | gold_answer | question | paragraph |
|---|---|---|---|---|
| Name of the document containing the paragraph | Type of question in the categorization of RAG tasks and challenges | The expected answer | Question posed to the model | Paragraph containing the question and answer |

The use of these two datasets allows for an initial and precise evaluation of the RAG system's performance in responding to factoid questions in Persian. This approach not only contributes to improving the quality of responses but also paves the way for future developments in the field of Persian natural language processing and intelligent question-answering systems.

## VI. CONCLUSION

In this paper, we have identified and addressed the key challenges faced when applying RAG in low-resource contexts. As a potential case study, we concentrated in developing a practical RAG system for Persian Language. Through our analysis, we proposed targeted solutions for addressing several issues containing table embedding, PDF files reading, chunking optimization, embedding model selection, retrieved chunks formatting, prompt engineering, questions answering in the target language, LLM selection, precise evaluation framework and datasets, hyperparameters optimization and, hallucination reduction.

Our experimental results show that the proposed solutions can significantly enhance RAG performance, making it a viable option for low-resource settings. These findings contribute to the broader adoption of retrieval-augmented models in areas with limited access to large-scale data and computational power. Moreover, the solutions proposed in this work offer practical insights for future research on optimizing RAG models in diverse, low resource domains, paving the way for more accessible and robust AI technologies globally.